# Une expérience de sémantique inférentielle

Farid Nouioua, Daniel Kayser

LIPN – Université Paris-Nord, 99 Avenue J.B.Clément 93430 Villetaneuse
(nouiouaf, dk)@lipn.univ-paris13.fr

**Résumé**

Nous développons un système qui doit être capable d'effectuer les mêmes inférences que le lecteur humain d'un constat d'accident de la route, et plus particulièrement de déterminer les causes apparentes de l'accident. Nous décrivons le cadre sémantique général dans lequel nous nous plaçons, les niveaux linguistiques et sémantiques de l'analyse, et les règles d'inférence utilisées par ce système.

**Abstract**

We are developing a system that aims to perform the same inferences as a human reader, on car-crash reports. More precisely, we expect it to determine the causes of the accident as they appear from the text. We describe the general semantic framework in which our study takes place, the linguistic and semantic levels of analysis, and the inference rules used by the system.

**Mots-clés :** sémantique inférentielle, raisonnement non monotone, causalité

## 1 Introduction

La très grande majorité des tentatives pour formaliser la sémantique des langues naturelles se base sur l'idée que les propositions de la langue composent des unités de sens faisant référence à des entités d'un « univers de discours », et que le sens d'une proposition réside dans l'ensemble des conditions sur cet univers qui la rendent vraie (voir parmi de nombreuses autres références Cherchia & McConnell-Ginet, 1990 ou Chambreuil, 1998).

Les avantages et les inconvénients de ces approches sémantiques sont bien connus : précision et généralité des outils de représentation du sens, mais faiblesse des opérations inférentielles dont elles peuvent rendre compte, au regard de ce qu'infèrent les lecteurs humains.

Ainsi, l'analyse sémantique du texte (1) ne permettra pas de répondre à une question, aussi évidente pour un lecteur humain, que : « le conducteur du véhicule B devait-il s'arrêter ? ».

(1)		Étant à l'arrêt au feu rouge, j'ai été percuté à l'arrière par le véhicule B, son conducteur n'ayant pas réussi à s'arrêter.

Conscients de cette faiblesse, les partisans d'une sémantique référentielle renvoient à une étape pragmatique la charge de répondre à ce type de questions. La pragmatique doit ainsi accomplir une tâche pratiquement infaisable, puisqu'elle doit résoudre toutes les questions laissées en suspens par les étapes précédentes.

Nous voulons construire un système sur des bases très différentes : au lieu de considérer les unités lexicales comme référant à des objets, et la mise en phrase de ces unités comme un processus compositionnel visant à établir des conditions de vérité, nos hypothèses sont :

-	Les unités lexicales évoquent des inférences. L'idée d'une sémantique procédurale n'est pas nouvelle (voir p.ex. (Small, 1981) ou le système inférentiel de (Rieger, 1975) sur les idées de Schank), mais elle semble avoir été abandonnée au profit de théories mathématiquement plus élégantes, dont nous dirons par euphémisme que l'adéquation au langage est incertaine.

-	Comprendre un texte, c'est essayer de satisfaire simultanément des contraintes émanant des différents niveaux de langue, ce qui ne peut se faire par une séquence de modules où chacun transmet au suivant les résultats acquis à son niveau.

Le succès de notre système s'évaluera par sa capacité à restituer, à partir d'un texte, les mêmes inférences qu'un lecteur standard.

Évidemment, il ne s'agit pas de faire table rase de l'existant, ni de prétendre construire en un temps forcément limité un système opérant sur n'importe quel domaine ! Bien que certaines contraintes du niveau *pragmatique* puissent par exemple rendre évident pour un lecteur humain qu'un mot a été employé à la place d'un autre, et donc modifier le résultat de l'étape *lexicale*, nous conservons l'architecture séquentielle de la plupart des systèmes de T.A.L., et limitons pour l'instant l'idée d'équilibre entre contraintes à un petit nombre de phénomènes de langue se situant dans un même niveau.

Nous exposons dans cet article l'état d'avancement d'un système travaillant sur un domaine restreint, mais non-trivial : celui des accidents de la route. Notre but est que le système donne pour ***cause***(s) de l'accident la ou les mêmes éléments qu'un lecteur ordinaire d'un constat amiable explicitant ses circonstances. Nous utilisons une conception de la causalité, développée dans (Kayser & Nouioua, 2005) et sur laquelle nous ne revenons pas ici, dans laquelle la cause est liée à la ***norme*** la plus spécifique qui a été violée.

Le plan de l'article est le suivant : la section 2 décrit les traitements qui élaborent, à partir du texte, des littéraux dans une logique du premier ordre réifié ne comportant que des prédicats jugés pertinents pour notre application. La section 3 montre comment ces prédicats déclenchent des inférences permettant d'extraire un « noyau », suffisant pour identifier ce qu'un lecteur humain considère comme étant la cause de l'accident. La section 4 met en perspective les résultats déjà obtenus et conclut l'article.

## 2   Les étapes « linguistiques »

### 2.1   Le corpus

Nous avons collecté auprès d'une compagnie d'assurances un nombre significatif de constats amiables rédigés après un accident de la route. Nous ne conservons que les rapports qui permettent à un lecteur standard, indépendamment du schéma qui les accompagne parfois, de déterminer ce qui, selon l'auteur du rapport, constitue la cause de l'accident[1]. La tâche à laquelle nous nous attaquons étant suffisamment ardue sans la compliquer davantage, nous avons rectifié quand il y avait lieu l'orthographe et/ou le temps grammatical de ces textes, mais dans notre conception du traitement de la langue, c'est le texte brut qui devrait être fourni, les contraintes orthographiques ayant le même statut que les autres contraintes, et les erreurs n'étant que des contraintes non satisfaites.

Nous travaillons sur un échantillon de 60 rapports[2] jugés représentatifs de ce corpus. Leur longueur varie de 9 à 167 mots. Ils totalisent 129 phrases dont la longueur va de 4 à 55 mots ; 24 rapports n'ont qu'une seule phrase, le plus long en a 7. L'échantillon comprend 2256 mots occurrences, mais seulement 500 mots distincts, provenant de 391 entrées de dictionnaire.

Nous utilisons TreeTagger[3] pour étiqueter ce corpus. Ce logiciel généraliste n'étant pas conçu pour les spécificités de nos textes, il commet quelques erreurs systématiques, dont rectifions les plus courantes par un post-traitement ; nous en profitons pour catégoriser plus finement quelques mots qui jouent par la suite un rôle important. En fin de compte, nous travaillons avec 41 catégories grammaticales. Le résultat de cette étape continue à comporter certaines erreurs ou imprécisions préjudiciables, mais que nous ne nous sommes pas permis de remettre en cause.

### 2.2   Niveau syntaxique

Contrairement au choix adopté pour l'étiquetage, nous avons créé un outil *ad hoc* pour l'analyse syntaxique de notre corpus. Quelques essais d'utilisation des logiciels existants nous ont montré que le post-traitement serait ici nettement plus important, et comme il nous semblait (l'expérience a montré que c'était à tort !) que les tournures utilisées dans ces rapports étaient assez peu variées, nous avons pensé que l'effort de conception et de réalisation d'un tel outil ne serait pas énorme. Une autre raison de notre choix est la nature des résultats attendus de l'analyse syntaxique : les éléments structurels qui sont cruciaux ici, et les critères pour les obtenir, ne sont pas nécessairement ceux que l'on retient dans l'analyse de textes quelconques.

Notre méthode consiste à découper manuellement chaque phrase du corpus en « blocs » : le bloc Pré englobe, lorsqu'il y en a, les circonstants antéposés ; PP contient la proposition principale, et le bloc Post, tout ce qui lui fait suite. La proposition principale est fragmentée en Groupe Nominal, Groupe Verbal, Complément Prépositionnel ; une grammaire de chacun de ces éléments est construite d'après le corpus, et on identifie ce qui, dans les blocs Pré et Post, est reconnu par ces grammaires locales ; une grammaire spécifique à chaque bloc est construite, toujours manuellement, en généralisant d'après les exemples.

---

[1] On notera que cette tâche est assez différente de celle traitée par (Després & Wolff, 02) sur un corpus de rapports de gendarmerie.
[2] l'exemple (1) de la section précédente en fait partie.
[3] http://www.ims.uni-stuttgart.de/projekte/corplex/TreeTagger/DecisionTreeTagger.html

Le découpage de (1) donne d'abord :

(2)     [Étant à l'arrêt au feu rouge]$_{Pré}$, [j'ai été percuté à l'arrière par le véhicule B]$_{PP}$ , [son conducteur n'ayant pas réussi à s'arrêter]$_{Post}$ .

Puis la proposition principale est découpée en :

(3)     [j']$_{GN}$ [ai été percuté]$_{GV}$ [à l'arrière]$_{CP}$ [par le véhicule B]$_{CP}$

Les « blocs » Pré et Post sont alors analysés de façon analogue. Cette méthode aboutit à une grammaire qui, en raison notamment des problèmes de rattachement, est intrinsèquement ambiguë. Sur des phrases longues, l'ambiguïté peut avoir des conséquences catastrophiques (plusieurs dizaines de milliers d'analyses proposées pour une même phrase). Nous avons donc essayé de la réduire au maximum, en mettant en facteur ce qui pouvait l'être sans inconvénient pour les relations syntaxiques que nous voulons extraire. Cet effort a réduit à un très petit nombre (mode < 8) les analyses d'une phrase ; cependant, 8 phrases sur 129 ont plus de 100 analyses, le maximum étant 900, ce qui dégrade la performance moyenne (32,4 analyses par phrase).

La grammaire utilise 100 non-terminaux, et comprend 340 règles de réécriture. Nous l'avons testée sur 21 nouveaux rapports, contenant 56 phrases. La performance d'ensemble reste stable (mode très proche de 7 ; moyenne : 32,8). Les temps d'exécution sont très rapides (réponse immédiate à l'échelle humaine), même lorsque le nombre d'analyses atteint le millier.

Le but de l'analyse est de tirer de chaque phrase un nombre limité de relations syntaxiques. L'avantage d'avoir écrit un analyseur ad hoc apparaît ici pleinement : il suffit en effet d'orner une minorité de règles (155 / 340) par un très petit nombre d'actions (34) pour que chaque analyse s'accompagne de la production des relations souhaitées.

Comme on pouvait l'espérer, de nombreuses analyses distinctes produisent le même ensemble de relations, ce qui diminue considérablement le travail des étapes suivantes. L'exemple (1) est un des plus délicats en raison de nombreuses ambiguïtés syntaxiques de rattachement ; c'est pourquoi le nombre d'analyses est ici très élevé (180) ; mais certaines analyses menant au même résultat, seuls 60 ensembles de relations distincts sont passés à l'étape suivante, qui en éliminera beaucoup sur critères d'attachement sémantique. Parmi ces ensembles, on trouve notamment (on ajoute des numéros à quelques mots pour distinguer leurs différentes occurrences dans le texte) :

*relation(PRÉ,être2,être1), support(être2,percuter), support(avoir,arrêter)*
*sujet(être1,je), sujet(être2,je), sujet(avoir,conducteur), objet(avoir,s'),*
*compl_v (à1,être1,arrêt), compl_v(à2,être2,arrière), compl_v(par,être2,véhicule),*
*compl_n(au,arrêt,feu), qualif_n(feu,rouge), qualif_n(véhicule,B), qualif_n(conducteur,son),*

*Une expérience de sémantique inférentielle*

*qualif_v(être1,PPRES), qualif_v(être2,PASSÉ), qualif_v(avoir,PPRES), qualif_v(avoir,NEG)*[4]

Ces résultats peuvent être aisément critiqués (les relations pourraient être affinées, elles regroupent des cas de figure syntaxiquement et sémantiquement différents, etc.). Mais nous n'avons aucun besoin d'une analyse syntaxique fine, et ce niveau d'analyse est suffisant pour les traitements qui vont suivre.

## 2.3   Construction des littéraux sémantiques

On infère maintenant, de l'ensemble de relations de surface obtenu ci-dessus, une représentation "sémantique" du texte et on réalise ainsi le passage de l'espace de représentation purement linguistique manipulant des mots du langage à un espace sémantique qui, lui, manipule des concepts. Il s'agit de transformer progressivement les relations de surface issues du texte en un ensemble de littéraux sémantiques exprimant les informations qu'il évoque ***explicitement***. Dans (Kayser & Nouioua, 2004), on trouve les détails du langage de représentation dont nous ne donnons ici que l'essentiel.

Ce langage du premier ordre « réifie » les noms de prédicats, c'est-à-dire qu'une relation binaire comme r(x,y) est représentée par le littéral *vrai(r,x,y)*. Les notions modales nécessaires à la détermination des causes, comme le fait pour un agent placé dans certaines circonstances de devoir effectuer une action, ou d'être en mesure de l'effectuer, s'exprimeront par *doit(agent, action, circonstance)* ou *en_mesure(agent, action, circonstance)*. Le rapport de l'accident décrit une séquence d'« états » (un état ne représente pas une situation « statique », mais un intervalle de temps durant lequel les prédicats jugés pertinents conservent une même valeur). L'argument *circonstance* des prédicats ci-dessus se réduit ainsi à l'identification d'un de ces intervalles.

Les littéraux du langage sont les prémisses de deux sortes de règles : les inférences strictes, traitées comme des implications, et les inférences par défaut au sens de (Reiter, 1980)[5]. Ces règles produisent des littéraux sémantiques, en réduisant à une même représentation logique des situations exprimées en langue par divers moyens lexicaux et/ou syntaxiques (synonymie, expressions anaphoriques, etc.).

**Factorisation du lexique**

L'intérêt des inférences vient essentiellement de leur généricité permettant à une règle de s'adapter à différentes situations. Pour assurer cette généricité, nous utilisons deux manières de regrouper des mots du lexique. La première consiste à rassembler dans une même classe d'équivalence les mots auxquels on attache en général le même sens dans le contexte de notre application. On utilise pour cela la relation *val_sem(X, Y)*, qui veut dire que *Y* est la valeur sémantique du mot *X* dans ce contexte. La deuxième manière permet de regrouper dans un même type des mots partageant un certain nombre de caractéristiques communes qui suffisent à déclencher des inférences : *type(X, Y)* exprime que *Y* est le type du mot *X*.

---

[4] Ces littéraux portent un nom explicitant la relation grammaticale trouvée (p.ex. qualif_n (_v) désigne une qualification du nom (du verbe) etc.). Nous employons « support » dans un sens élargi par rapport à Gaston Gross (Gross, 99) : un support et son verbe sont présumés former un seul élément prédicatif, donc une relation peut s'adresser indifféremment au verbe ou au support.

[5] La notation A : B [C] représente le défaut A : B∧C / B qui exprime le fait que si A a été déduit, B le sera aussi sauf preuve du contraire (i.e. ¬B est déductible) ou blocage de cette règle (cas où ¬C est déductible).

**Prétraitements des relations de surface**

Des traitements préalables, procéduraux et inférentiels, sont effectués afin d'extraire des relations de surface ce qui est nécessaire aux processus d'inférence des littéraux sémantiques.

La résolution des anaphores est un besoin qui s'impose dans les textes du corpus. Il s'agit d'un problème difficile dans le cas général. Cependant la limitation de notre champ d'étude nous facilite la tâche. On tire profit des avancées réalisées dans cet axe de recherche (Mitkov, 2002) pour adapter une heuristique à notre contexte. L'algorithme doit produire la relation *même_ref(X, Y)* pour exprimer que « *X et Y réfèrent la même entité* ». Dans notre exemple, on obtiendra : *même_ref(j', Auteur), même_ref(son, véhicule)*. Ce dernier résultat aboutira à ce que la relation *qualif_n(conducteur,son)* soit équivalente à une qualification conducteur du véhicule B. De plus, la métonymie très répandue qui assimile un conducteur à son véhicule se traduit par la règle : «*si on a une relation qualif-n(X,Y) où X est de type agent (type(X,agent)) et Y est un symbole (A, B, C, ..), remplacer par Y toutes les occurrences de X ainsi que les Z ayant la même référence que X (même_ref(X,Z))*». Une première application de cette règle remplace véhicule et son par B, et une deuxième application de la même règle remplace conducteur par B.

Quelques expressions linguistiques courantes sont reconnues au préalable : ainsi, sur notre exemple, chaque occurrence de l'expression feu rouge est remplacée par l'entité « *feu_rouge* ». D'autres prétraitements détectent et gèrent la voie passive. Enfin, nous procédons au remplacement du verbe support par le verbe supporté ; ainsi, dans notre exemple le verbe avoir est remplacé par le verbe arrêter[6].

**Production de littéraux sémantiques « intermédiaires »**

Le passage par un niveau intermédiaire est motivé par une raison méthodologique : il facilite l'écriture des règles d'inférence qui suivent en factorisant leurs prémisses et permet ainsi une meilleure lisibilité. À ce niveau, on fait apparaître également l'information temporelle dans les littéraux. Cette information est centrée sur les verbes et les prépositions, et figure en tant que paramètre dans un littéral intermédiaire sous la forme : *ref_temp(X)* qui indique "la référence temporelle liée au mot *X*". Examinons quelques exemples de règles du niveau intermédiaire :

Le défaut (4) traite des verbes transitifs en regroupant dans une seule forme, plus proche de la forme finale voulue, le verbe, son sujet et son objet :

(4)     *sujet(X, Y) ∧ objet(X, Z) : vrai(combine(X, Z), Y, ref_temp(X))[¬qualif_v(X, neg)]*

ce qui signifie : si le sujet d'un verbe *X* est *Y*, et son complément d'objet *Z*, déduire que *Y* vérifie le prédicat complexe (*X,Z*) au temps référencé par *X*, sauf si *X* est à une forme négative.

Des règles analogues permettent de traiter les cas des verbes intransitifs et des compléments prépositionnels. Voici les littéraux sémantiques intermédiaires obtenus pour la phrase (1) :

---

[6] Une liste de verbes supports est établie à priori. Des traitements plus compliqués peuvent être envisagés. Par exemple, le support oublier introduit une négation sur le verbe supporté ; de même pour le support empêcher qui en plus affecte son objet comme sujet au verbe supporté…

*vrai(être1,Auteur,ref_temp(être1))*, *vrai(percuter,Auteur,ref_temp(percuter))*,
¬*vrai(arrêter,B,ref_temp(arrêter))*, *vrai(combine(arrêter,se),B,ref_temp(arrêter)*,
*vrai(combine(combine(à1,être1),arrêt),Auteur,ref_temp(à1))*,
*vrai(combine(combine(à2,percuter),arrière),Auteur,ref_temp(à2))*,
*vrai(combine(combine(par,percuter),B),Auteur,ref_temp(par))*.

**Détermination des noms des littéraux sémantiques, et calcul des relations temporelles**

Les règles de cette étape permettent d'une part de fixer définitivement les arguments des littéraux sémantiques, et d'autre part de générer un ensemble de relations entre les références temporelles. Deux types de relations sont considérées : *Préc(T1, T2)* qui veut dire que *T1* précède *T2* et *Simul(T1, T2)* qui exprime le fait que *T1* et *T2* sont simultanés. Considérons le fonctionnement de quelques règles sur notre exemple :

(5)   *type(L,agent)* ∧ *val_sem(X,à)* ∧ *val_sem(V,être)* ∧ *val_sem(N,arrêt)* ∧
*vrai(V,L,ref_temp(V))* ∧ *vrai(combine(combine(X,V),N),L,ref_temp(X))* : *vrai(arrêter, L, ref_temp(V))*

Cette règle permet d'inférer (avec *L = Auteur*) à partir de *"Etant à l'arrêt..., j'..."*, le littéral : *vrai(arrêter, Auteur, ref_temp(être))*. Tandis que la règle (6) infère (avec *L = Auteur, Z = à2, V = percuter, N = arrière*) les deux littéraux : *vrai(combine(position_choc, arrière), Auteur, ref_temp(à2))* et *simul(ref_temp(percuter),ref_temp(à2))*.

(6)   *prep(Z)* ∧ *type(L,véhicule)* ∧ *type(N,position)* ∧ *val_sem(V,heurter)* ∧ *vrai(combine(combine(Z, V), N), L, ref_temp(Z))* ∧ *voie(V, passive)* : *vrai(combine(position_choc, arrière), L, ref_temp(V))* ∧ *simul(ref_temp(V), ref_temp(Z))*

Des règles de même nature infèrent le reste des littéraux.

**Calcul des paramètres temporels : produire la version finale des littéraux sémantiques**

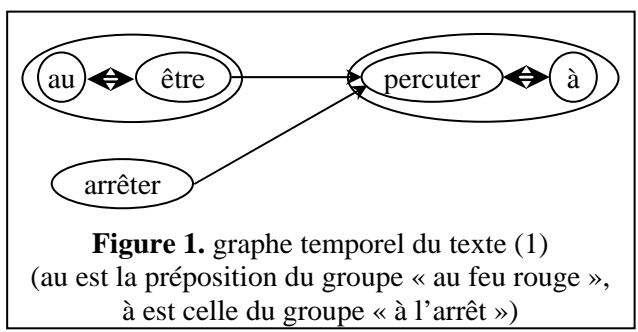

**Figure 1.** graphe temporel du texte (1)
(au est la préposition du groupe « au feu rouge »,
à est celle du groupe « à l'arrêt »)

La dernière tâche à effectuer avant de trouver la version finale des prédicats sémantiques consiste à ordonner les références temporelles. Une méthode générale pour ce faire est de modéliser les relations *préc* par un graphe orienté *G* dont les nœuds sont les références temporelles. *(T1, T2)* est un arc de *G* si et seulement si on dispose de la relation *préc(T1, T2)*. Le graphe obtenu est, bien entendu, acyclique (il est impossible de trouver deux instants distincts où chacun est situé avant l'autre !). La figure 1. représente le graphe correspondant à notre texte. On décompose le graphe *G* en niveaux. Les références temporelles appartenant au $i^{ème}$ niveau du graphe sont remplacées par la valeur entière *i*. Enfin, on utilise la relation *simul* pour attribuer les bonnes valeurs aux références temporelles restantes. Dans notre exemple, la décomposition du graphe permet d'obtenir: *ref_temp(être) = ref_temp(arrêter) =1; ref_temp(percuter) = 2*. Les relations : *simul(ref_temp(être), ref_temp(au))* et *simul(ref_temp(percuter), ref_temp(à))* permettent d'en déduire : *ref_temp(au) = 1* et *ref_temp(à) = 2*. La liste des littéraux sémantiques obtenus sera ainsi : *vrai(arrêter,Auteur,1)*,

*vrai(feu_rouge,Auteur,1), ¬vrai(arrêter,B,1), vrai(combine(heurter,Auteur),B,2), vrai(combine(position_choc, arrière), Auteur,2).*

# 3 Les étapes « sémantiques »

Le raisonnement sémantique que nous devons entreprendre sur la représentation ainsi obtenue a pour but de trouver la cause de l'accident. Les règles employées sont basées sur nos connaissances sur les normes du domaine de la route et permettent d'enrichir les littéraux sémantiques résultant des traitements linguistiques par de nouveaux littéraux incorporant des informations *implicites* sur l'accident. Bien que le texte ne les énonce pas explicitement, ces informations sont indispensables à sa compréhension. Le processus de raisonnement sémantique converge vers un noyau[7] d'un nombre restreint de littéraux exprimant des sens élémentaires, qui suffisent pour exprimer les différentes causes possibles exprimées dans les réponses d'un lecteur humain ordinaire.

La cause recherchée correspond à une violation d'une norme du domaine appelée aussi anomalie. Si plusieurs anomalies sont présentes, on distingue la *« vraie anomalie »,* qui explique le mieux l'accident, des anomalies qui en résultent. Une vraie anomalie est détectée notamment si « à un instant *T*, un agent *A* doit atteindre un effet[8] *E*, qu'il est en mesure de le faire mais qu'à l'instant *T+1*, un effet *E'* incompatible avec *E* est vérifié » (règle (7)) voir (Kayser, Nouioua 2004) (Kayser, Nouioua 2005) pour plus de détails).

(7) $doit(E,A,T) \land en\_mesure(E,A,T) \land vrai(E',A,T+1) \land incompatible(E,E') \rightarrow vraie\_An$

## 3.1 Réduction des littéraux sémantiques aux littéraux noyaux

Pour illustrer la nature des inférences de ce niveau, prenons deux exemples de règles : la règle stricte (8) exprime le fait que : *"Si W heurte V à un instant T, alors W n'est pas à l'arrêt à cet instant T"*. Elle permet de déduire le littéral : ¬*vrai(arrêter, B, 2)*. Le défaut (9) exprime le fait que : *"En général, s'il y a un choc entre V et W à l'instant T et que la position du choc de V est l'arrière, alors W était le suivant de V dans une file à l'instant T-1. Cette inférence est bloquée si W n'a pas le contrôle à l'instant T-1"*. Ce défaut permet d'inférer le littéral : *vrai(combine(suiv, Auteur), B, 1)*.

Des règles de même nature infèrent les prédicats : *vrai(combine(choc, Auteur), B, 2), vrai(combine(choc, B), Auteur, T), vrai(combine(obstacle, Auteur), B, 1), vrai(combine(obstacle, B), Auteur, 1).*

(8) $vrai(combine(heurter, V), W, T) \rightarrow \neg vrai(arrêter, W, T)$

(9) $vrai(combine(choc, V), W, T) \land vrai(combine(position\_choc, arrière), V, T) :$
    $vrai(combine(suiv, V), W, T-1) [vrai(contrôle, W, T-1)]$

---

[7] Les prédicats du noyau concernent les propriétés : arrêt, contrôle, démarrer, reculer, rouler_lentement et combine(cause_perturbation_anormale, X) i.e. X est une cause de perturbation anormale

[8] Nous utilisons une « ontologie » qui distingue a priori des noms d'action Act (ce qu'un agent peut volontairement décider d'entreprendre) et des noms d'effet E (le résultat anticipé d'une action) ; la connexion entre les deux s'effectue par la relation raison_pot(Act,E) (vouloir obtenir E est une raison potentielle pour entreprendre Act).

## 3.2 Détection de la cause à partir du noyau

Une première règle, exprimant le fait que *"On doit à tout instant éviter les obstacles s'ils se présentent"*, permet d'inférer le devoir de chacun des agents Auteur et B d'éviter l'autre. Le défaut (10) exprime que le meilleur moyen d'éviter un obstacle est de s'arrêter (sauf si l'obstacle nous suit, ou si on est déjà à l'arrêt, ou quelques autres cas). Son application permet d'inférer *doit(arrêt, B, 1)*. Ce devoir n'est pas inféré pour *Auteur* car le défaut dans ce cas est bloqué par *vrai(combine(suiv, Auteur), B, 1)*.

(10)   *doit(combine(éviter, W), V, T) ∧ vrai(combine(choc, W), V, T+1) : doit(arrêt, V, T)*
 *[¬ vrai(combine(suiv, V), W, T), ¬vrai(arrêt, V, T), ¬doit(roule_lentement, V, T),*
 *¬doit(non(reculer), V, T-1), ¬doit(non(démarrer), V, T-1),*
 *prévisible(combine(obstacle, W), V, T) ]*

La règle (11) permet d'inférer la capacité d'un agent. Sa signification est que : *"V est en mesure à l'instant T d'atteindre un effet E, si et seulement s'il existe une action Act qui est une raison potentielle pour E et qui est disponible pour V à l'instant T"* :

(11)   *en_mesure(E,V,T) ↔ action(Act) ∧ effet(E) ∧ raison_pot(Act, E) ∧ disponible(Act,E,V,T)*

L'ensemble des actions, des effets et des raisons potentielles sont prédéterminés statiquement, et on a par exemple : *raison_pot(freiner,arrêter)*.

Par défaut, les actions sont disponibles pour les agents. Ce défaut est inhibé en présence d'un certain nombre d'exceptions (absence de contrôle, présence d'un événement non maîtrisable, ...). Aucune de ces exceptions n'est présente dans notre cas. Le défaut s'applique donc et donne : *disponible(freiner, arrêter,B,1)*. Ce qui nous permet d'inférer : *en_mesure(arrêter, B, 1)*.

Les littéraux : *doit(arrêter, B, 1)*, *en_mesure(arrêter, B, 1)* et *vrai(arrêter, B, 2)* déclenchent la première forme d'une vraie anomalie (7) et déduisent : *Vraie_An*, ce qui met un terme à la recherche de la cause. Celle-ci est exprimée comme suit :

*" Le véhicule B avait à l'instant 1 le devoir de s'arrêter afin d'éviter le choc, mais il n'a pas respecté son devoir car à l'instant 2, il n'était pas à l'arrêt ".*

## 4 Discussion, conclusion et perspectives

Le système est en cours d'implémentation. Nous avons utilisé pour cela, le langage SMODELS[9] basé sur la théorie des modèles stables (Gelfond, Lifschitz, 2004). Chaque modèle stable trouvé par SMODELS est formé de l'ensemble des littéraux appartenant à une extension de la théorie des défauts sous-jacente. La partie du raisonnement qui cherche la cause de l'accident à partir des littéraux sémantiques est complètement testée. Avec environ 200 règles, le système trouve, pour chacun de nos 60 textes, un seul modèle stable contenant les littéraux nécessaires à l'expression de la cause de l'accident. Le temps d'exécution varie d'un texte à l'autre, mais ne dépasse pas 30 secondes. Comme nous avons « traité à la main » les 60 rapports de l'échantillon en supposant résolues les étapes menant aux littéraux sémantiques, nous savons que les causes trouvées par notre système seront bien celles que nous espérions obtenir, dès que les dernières difficultés linguistiques seront résolues. Il est évident qu'elles le seront pour notre échantillon, au prix d'un traitement ad hoc s'il le faut.

---

[9] *SMODELS* travaille sur le résultat d'un autre logiciel : *LPARSE* qui entre autres instancie les variables afin de produire des règles propositionnelles. SMODELS et LPARSE sont disponibles sur le Web à l'adresse: http://www.tcs.hut.fi/Software/smodels/

Il est non moins évident qu'un ajout de nouveaux rapports dégradera les performances de notre système, ne serait-ce qu'en raison de mots et de tournures qui ne figurent pas dans l'échantillon. Ce qui est difficilement évaluable actuellement, c'est l'ampleur de cette dégradation, et surtout l'effort à fournir pour y remédier. Notre grammaire est assez fragile, et il faudra peut-être recourir à des analyseurs généralistes, surtout si ceux-ci s'améliorent. Nous pensons au contraire que les autres étapes : la construction des littéraux sémantiques, leur réduction à un noyau, et l'inférence de la cause, sont assez robustes. Cette conjecture ne tardera pas à être vérifiée ou infirmée.

Supposons qu'elle soit vérifiée : pour que la valeur de notre travail dépasse la simple prouesse d'ingéniosité, il faudra utiliser la même approche dans un autre domaine. Ce n'est que lorsqu'elle aura fait ses preuves, en fournissant des inférences analogues à celles des humains, dans différentes spécialités que l'on pourra envisager d'élargir le champ d'application, et collecter une fraction de plus en plus importante des normes qui guident le raisonnement de sens commun. Cette démarche nous semble plus raisonnable que celle consistant, comme CYC (Lenat & Guha, 1990), à attaquer de front l'ensemble des connaissances humaines.

Simultanément, il serait intéressant de faire interagir un plus grand réseau de contraintes, afin que les extensions trouvées par le système de raisonnement (SMODELS ou un autre) prennent en compte les défauts utilisés à tous les niveaux. Ceci permettrait d'aboutir pour chaque texte à l'interprétation qui satisfait simultanément le mieux l'ensemble des contraintes identifiées, qu'elles soient orthographiques, morpho-lexicales, syntaxiques, ou sémantico-pragmatiques.

*Une expérience de sémantique inférentielle*